\documentclass[sigconf]{acmart}

\setcopyright{rightsretained}

\copyrightyear{2020}
\acmYear{2020}
\acmConference[WWW '20]{Proceedings of The Web Conference 2020}{April 20--24,2020}{Taipei, Taiwan}
\acmBooktitle{Proceedings of The Web Conference 2020 (WWW '20), April 20--24, 2020, Taipei, Taiwan}
\acmPrice{}
\acmDOI{10.1145/3366423.3380132}
\acmISBN{978-1-4503-7023-3/20/04}

\usepackage{amsmath, amsfonts, amssymb, amsthm, xspace, color}
\usepackage{booktabs} 

\usepackage{subfigure, multirow, tabularx} 
\usepackage{booktabs} 
\usepackage{enumitem}
\usepackage{flushend}
\usepackage[T1]{fontenc}
\usepackage{epstopdf}
\usepackage{balance}
\usepackage{graphicx}
\usepackage[noend,ruled,linesnumbered]{algorithm2e} 
\usepackage{microtype}
\usepackage{url}
\usepackage{wrapfig,lipsum}
\usepackage{makecell}
\usepackage{wrapfig}

\newcommand{\hide}[1]{} 

\newcommand{\ie}{\emph{i.e.}\xspace} 
\newcommand{\eg}{\emph{e.g.}\xspace} 
\newcommand{\nop}[1]{}
\newcommand{\mquote}[1]{{``\emph{#1}''}}

\newcommand{\naive}{na\"{\i}ve\xspace} 

\newtheorem{thm:def}{Definition}
\newtheorem{thm:eg}{Example}
\newtheorem{thm:lem}{Lemma}
\newtheorem{thm:obs}{Observation}
\newtheorem{thm:req}{Requirement}
\newtheorem{thm:prop}{Proposition}
\newtheorem{thm:principle}{Principle}
\newtheorem{thm:thm}{Theorem}
\newtheorem{thm:corollary}{Corollary}

\newcommand{\wtd}[1]{\widetilde{#1}}			
\newcommand{\pair}[1]{\langle #1 \rangle}			

\DeclareMathOperator*{\argmax}{arg\,max}

\DeclareMathOperator*{\concat}{\scalebox{1}[2.5]{$\parallel$}}

\def \P {\mathbf{P}}

\def \C {\mathcal{C}}
\def \D {\mathcal{D}}
\def \E {\mathcal{E}}

\def \N {\mathcal{N}}

\def \R {\mathcal{R}}

\def \T {\mathcal{T}}

\newcommand{\TaxoExpan}{\mbox{\sf TaxoExpan}\xspace}
\newcommand{\TaxoExpanBf}{\mbox{\sf \textbf{TaxoExpan}}\xspace}
\newcommand{\TaxoExpanFWFS}{\mbox{\sf TaxoExpan-FWFS}\xspace}
\newcommand{\TaxoExpanFWFSBf}{\mbox{\sf \textbf{TaxoExpan-FWFS}}\xspace}

\definecolor{midnightgreen}{rgb}{0.0, 0.29, 0.33}
\definecolor{orange}{RGB}{255,127,0}

\begin{document}

\fancyhead{}  

\title{TaxoExpan: Self-supervised Taxonomy Expansion with Position-Enhanced Graph Neural Network}

\author{Jiaming Shen$^{1\star}$, Zhihong Shen$^{2}$, Chenyan Xiong$^{2}$, Chi Wang$^{2}$, Kuansan Wang$^{2}$, Jiawei Han$^{1}$}
\affiliation{%
  \institution{$^1$University of Illinois at Urbana-Champaign, IL, USA $\quad$ $^2$ Microsoft Research, WA, USA}
}
\affiliation{%
  \institution{$^1$\{js2, hanj\}@illinois.edu $\quad$ $^2$\{Zhihosh, Chenyan.xiong, Wang.chi, kuansanw\}@microsoft.com}
}

\renewcommand{\shorttitle}{TaxoExpan}
\renewcommand{\shortauthors}{J. Shen et al.}

\begin{abstract}

Taxonomies consist of machine-interpretable semantics and provide valuable knowledge for many web applications. 
For example, online retailers (\eg, \texttt{Amazon} and \texttt{eBay}) use taxonomies for product recommendation, and web search engines (\eg, \texttt{Google} and \texttt{Bing}) leverage taxonomies to enhance query understanding.
Enormous efforts have been made on constructing taxonomies either manually or semi-automatically.
However, with the fast-growing volume of web content, existing taxonomies will become outdated and fail to capture emerging knowledge.
Therefore, in many applications, dynamic expansions of an existing taxonomy are in great demand. 
In this paper, we study how to expand an existing taxonomy by adding a set of new concepts.
We propose a novel self-supervised framework, named \TaxoExpan, which automatically generates a set of $\langle$query concept, anchor concept$\rangle$ pairs from the existing taxonomy as training data.
Using such \emph{self-supervision} data, \TaxoExpan learns a model to predict whether a query concept is the direct hyponym of an anchor concept. 
We develop two innovative techniques in \TaxoExpan: 
(1) a position-enhanced graph neural network that encodes the local structure of an anchor concept in the existing taxonomy, and 
(2) a noise-robust training objective that enables the learned model to be insensitive to the label noise in the self-supervision data. 
Extensive experiments on three large-scale datasets from different domains demonstrate both the effectiveness and the efficiency of \TaxoExpan for taxonomy expansion. 

\end{abstract}


\maketitle

{
\renewcommand{\thefootnote}{\fnsymbol{footnote}}
\footnotetext[1]{This work is done while interning at Microsoft Research.}
}

\section{Introduction}\label{sec:intro}

Taxonomies have been fundamental to organizing knowledge for centuries \cite{Stewart2008BuildingET}.
In today's Web, taxonomies provide valuable knowledge to support many applications such as query understanding~\cite{Hua2017UnderstandST}, content browsing~\cite{Yang2012ConstructingTT}, personalized recommendation~\cite{Zhang2014TaxonomyDF, Huang2019TaxonomyAwareMR}, and web search~\cite{Wu2012ProbaseAP, Liu2019AUC}. 
For example, many online retailers (\eg, \texttt{eBay} and \texttt{Amazon}) organize products into categories of different granularities, so that customers can easily search and navigate this category taxonomy to find the items they want to purchase. 
In addition, web search engines (\eg, \texttt{Google} and \texttt{Bing}) leverage a taxonomy to better understand user queries and improve the search quality. 

Existing taxonomies are mostly constructed by human experts or in a crowdsourcing manner.  
Such manual curations are time-consuming, labor-intensive, and rarely complete. 
To reduce the human efforts, many automatic taxonomy construction methods~\cite{Shen2018HiExpanTT, Zhang2018TaxoGenCT, Mao2018EndtoEndRL} are proposed. 
They first identify ``is-A'' relations (\eg, \mquote{iPad} is an \mquote{Electronics}) using textual patterns~\cite{Hearst1992AutomaticAO, Roller2018HearstPR} or distributional similarities~\cite{shwartz2016improving, Anke2016SupervisedDH}, and then organize extracted concept pairs into a directed acyclic graph (DAG) as the output taxonomy~\cite{Kozareva2010ASM, Gupta2017TaxonomyIU, Cocos2018ComparingCF}. 
As the web contents and human knowledge are constantly growing, people need to expand an existing taxonomy to include new emerging concepts. 
Most of previous methods, however, construct a taxonomy entirely \emph{from scratch} and thus when we add new concepts, we have to re-run the entire taxonomy construction process. 
Although being intuitive, this approach has several limitations. 
First, many taxonomies have a top-level design provided by domain experts and such design shall be preserved.
Second, a newly constructed taxonomy may not be consistent with the old one, which can lead to instabilities of its dependent downstream applications. 
Finally, as targeting the scenario of building taxonomy from scratch, most previous methods are unsupervised and cannot leverage signals from the existing taxonomy to construct a new one. 

\begin{figure}[!t]
  \centering
  \centerline{\includegraphics[width=0.48\textwidth]{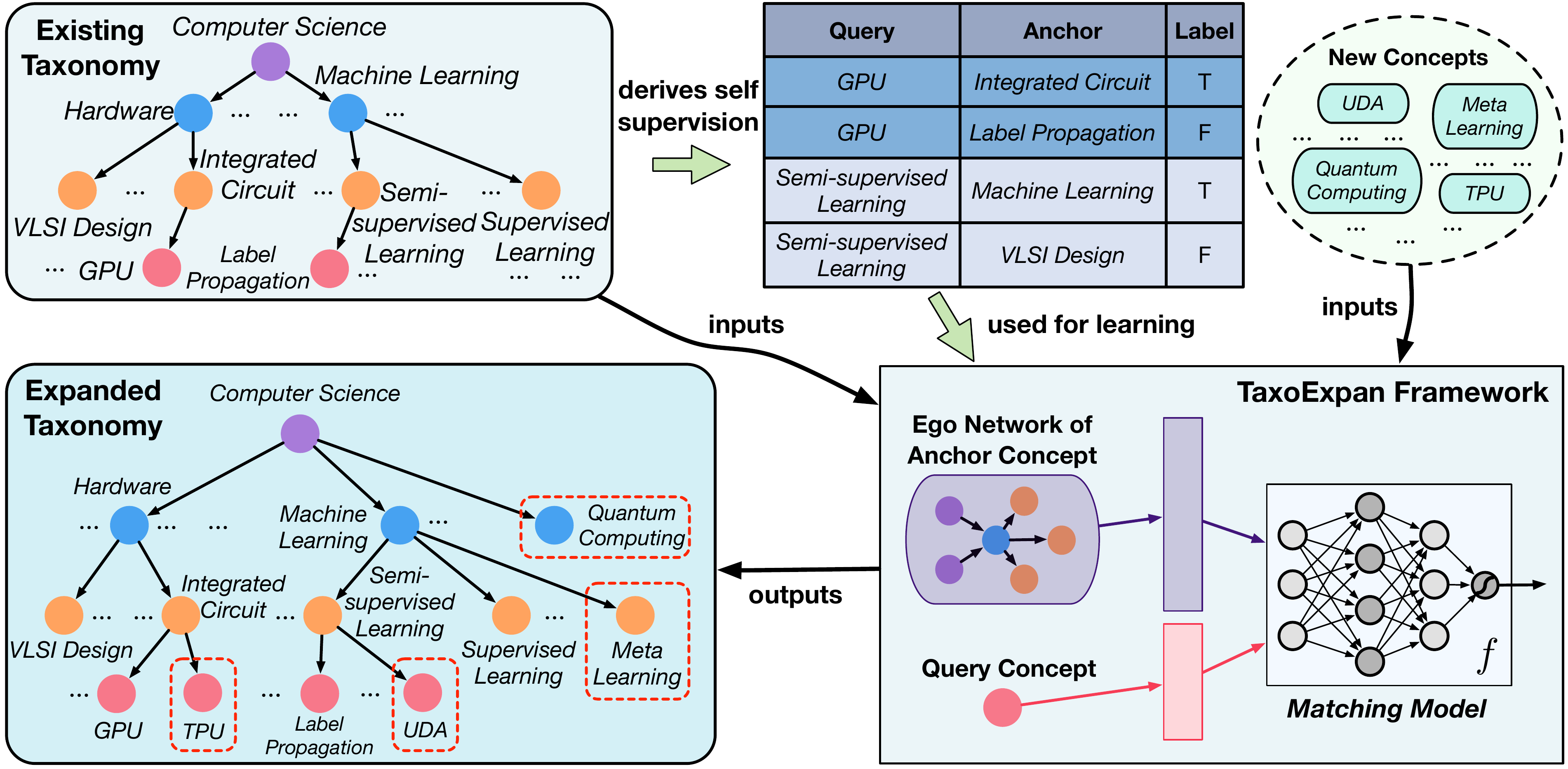}}
  \vspace{-0.2cm}
  \caption{An example of expanding one computer science field-of-study taxonomy to include new concepts such as \mquote{Quantum Computing}, \mquote{Meta Learning}, and \mquote{TPU}.}
  \label{fig:task}
  \vspace{-0.3cm}
\end{figure}

In this paper, we study the \textit{taxonomy expansion} task: given an existing taxonomy and a set of new emerging concepts, we aim to automatically expand the taxonomy to incorporate these new concepts (without changing the existing relations in the given taxonomy).\footnote{\scriptsize We recognize that the modification of an existing taxonomy is necessary in some cases. However, it happens much less frequently and requires high cautiousness from human curator. Therefore, we leave it out of the scope of automation.}
Figure~\ref{fig:task} shows an example where a taxonomy in computer science domain is expanded to include new subfields (\eg, \mquote{Quantum Computing}) and new techniques (\eg, \mquote{Meta Learning} and \mquote{UDA}). 
Some previous studies \cite{Jurgens2015ReseratingTA, Jurgens2016SemEval2016T1, Schlichtkrull2016MSejrKuAS} attempt this task by using an additional set of labeled concepts with their true insertion positions in the existing taxonomy.
However, such labeled data are usually small and thus forbid us from learning a more powerful model that captures the subsumption semantics in the existing taxonomy.  


We propose a novel framework named \TaxoExpan to tackle the lack-of-supervision challenge.
\TaxoExpan formulates a taxonomy as a directed acyclic graph (DAG), automatically generates pseudo-training data from the existing taxonomy, and uses them to learn a matching model for expanding a given taxonomy.
Specifically, we view each concept in the existing taxonomy as a \emph{query} and one of its parent concepts as an \emph{anchor}.
This gives us a set of positive $\pair{\text{query concept, anchor concept}}$ pairs. 
Then, we generate negative pairs by sampling those concepts that are neither the descendants nor the direct parents of the query concept in the existing taxonomy. 
In Figure~\ref{fig:task}, for example, the $\langle$\mquote{GPU}, \mquote{Integrated Circuit}$\rangle$ is a positive pair and $\langle$\mquote{GPU}, \mquote{Label Propagation}$\rangle$ is a negative pair. 
We refer to these training pairs as \emph{self-supervision} data, because they are procedurally generated from the existing taxonomy  and no human curation is involved. 

To make the best use of above self-supervision data, we develop two novel techniques in \TaxoExpan. 
The first one is a position-enhanced graph neural network (GNN) which encodes the local structure of an anchor concept using its ego network (egonet) in the existing taxonomy. 
If we view this anchor concept as the ``parent'' of the query concept, this ego network includes the potential ``siblings'' and ``grand parents'' of the query concept. 
We apply graph neural networks (GNNs) to model this ego network. 
However, regular GNNs fail to distinguish nodes with different relative positions to the query (\ie, some nodes are grand parents of the query while the others are siblings of the query).
To address this limitation, we present a simple but effective enhancement to inject such position information into GNNs using position embedding.
We show that such embedding can be easily integrated with existing GNN architectures (\eg, GCN \cite{Kipf2017SemiSupervisedCW} and GAT \cite{Velickovic2018GraphAN}) and significantly boosts the prediction performance. 
The second technique is a new noise-robust training scheme based on the InfoNCE loss \cite{Oord2018RepresentationLW}. 
Instead of predicting whether each individual $\pair{\text{query concept, anchor concept}}$ pair is positive or not, we first group all pairs sharing the same query concept into a single training instance and learn a model to select the positive pair among other negative ones from the group. 
We show that such training scheme is robust to the label noise and leads to performance gains. 

We test the effectiveness of \TaxoExpan framework on three real-world taxonomies from different domains. 
Our results show that \TaxoExpan can generate high-quality concept taxonomies in scientific domains and achieves state-of-the-art performance on the WordNet taxonomy expansion challenge \cite{Jurgens2016SemEval2016T1}.  

\smallskip
\noindent \textbf{Contributions.}
To summarize, our major contributions include:
(1) a self-supervised framework that automatically expands existing taxonomies without manually labeled data; 
(2) an effective method for enhancing graph neural network by incorporating hierarchical positional information; 
(3) a new training objective that enables the learned model to be robust to label noises in self-supervision data; and
(4) extensive experiments that verify both the effectiveness and the efficiency of \TaxoExpan framework on three real-world large-scale taxonomies from different domains.

\smallskip
The rest of the paper is organized as follows.
Section~\ref{sec:related_work} discusses the related work.
Section~\ref{sec:problem} formalizes our problem.
Then, we present our \TaxoExpan framework in Section~\ref{sec:method} and conduct experiments in Section~\ref{sec:exp}.
Finally, we conclude this paper in Section~\ref{sec:conclusion}.

\section{Related Work}\label{sec:related_work}
We review two lines of related work: taxonomy construction and graph neural network. 

\smallskip
\noindent \textbf{Taxonomy Construction and Expansion.}
Automatic taxonomy construction is a long-standing task in the literature.
Most existing approaches focus on building the \emph{entire} taxonomy by first extracting hypernym-hyponym pairs and then organizing all hypernymy relations into a tree or DAG structure. 
For the first hypernymy discovery step, methods fall into two categories: (1) \emph{pattern-based} methods which leverage pre-defined patterns \cite{Hearst1992AutomaticAO, Agichtein2000SnowballER, Nakashole2012PATTYAT, Jiang2017MetaPADMP} to extract hypernymy relations from a corpus, and (2) \emph{distributional} methods which calculate pairwise term similarity metrics based on term embeddings \cite{Lin1998AnID, Weeds2004CharacterisingMO, Roller2014InclusiveYS, Luu2016LearningTE} and use them to predict whether two terms hold the hypernymy relation. 
For the second hypernymy organization step, most methods formulate it as a graph optimization problem.
They first build a noisy hypernymy graph using hypernymy pairs extracted and then derive the output taxonomy as a particular tree or DAG structure (\eg, maximum spanning tree \cite{Navigli2011AGA, Bansal2014StructuredLF}, optimal branching \cite{Velardi2013OntoLearnRA}, and minimum-cost flow \cite{Gupta2017TaxonomyIU}) from the hypernymy graph. 
Finally, there are some methods that leverage entity set expansion techniques~\cite{Zhang2017EntitySE, Shen2017SetExpanCS} to incrementally construct a taxonomy either from scratch or from a tiny seed taxonomy.

In many real-world applications, some existing taxonomies may have already been laboriously curated by experts \cite{fellbaum1998wordnet, Lipscomb2000MedicalSH} or via crowdsourcing \cite{Meng2015CrowdTCCT}, and are deployed in online systems. 
Instead of constructing the entire taxonomy from scratch, these applications demand the feature of expanding an existing taxonomy dynamically. 
There exist some studies on expanding WordNet with named entities from Wikipedia \cite{Toral2008NamedEW} or domain-specific concepts from different corpora \cite{Bentivogli2003ArchiWordNetIW, Fellbaum2006TowardsNI, Jurgens2015ReseratingTA, Anke2016ExtendingWW}.
Task 14 of SemEval 2016 challenge \cite{Jurgens2016SemEval2016T1} is specifically setup to enrich WordNet with concepts from domains like health, sport, and finance. 
One limitation of these approaches is that they depend on the synset structure unique to WordNet and thus cannot be easily generalized to other taxonomies. 

To address the above limitation, more recent works try to develop methodologies for expanding a generic taxonomy. 
Wang \textit{et al.} \cite{Wang2014AHD} design a hierarchical Dirichlet model to extend the category taxonomy in search engines using query logs. 
Plachouras \textit{et al.} \cite{Plachouras2018ACO} learn paraphrase models on external paraphrase datasets and apply learned models to directly find paraphrases of concepts in the existing taxonomy. 
Vedula \textit{et al.} \cite{Vedula2018EnrichingTW} combine multiple features, some of which are retrieved from an external \texttt{Bing} Search API, into a ranking model to score candidate positions in terms of their matching scores with the query concept. 
Aly \textit{et al.} \cite{Aly2019EveryCS} first learn term embeddings in a hyperbolic space and then attach each new concept to its most similar node in the existing taxonomy based on the hyperbolic embeddings.
Comparing with these methods, our \TaxoExpan framework has two advantages. 
First, it requires no external data resource and makes full use of the existing taxonomy as the self supervision, which leads to a broader application scope.
Second, \TaxoExpan explicitly models the local structure around each candidate position, which boosts the quality of expanded taxonomy. 

\smallskip
\noindent \textbf{Graph Neural Network.}
Our work is also related to Graph Neural Network (GNN) which is a generic method of learning on graph-structure data.
Many GNN architectures have been proposed to either learn individual node embeddings \cite{Kipf2017SemiSupervisedCW, Hamilton2017InductiveRL, Chen2018FastGCNFL, Velickovic2018GraphAN} for the node classification and the link prediction tasks or learn an entire graph representation \cite{Ying2018HierarchicalGR, Zhang2018AnED, Lee2018GraphCU} for the graph classification task.  
In this work, we tackle the taxonomy expansion task with a fundamentally different formulation from previous tasks.
We leverage some existing GNN architectures and enrich them with additional relative position information. 
Recently, You \textit{et al.} \cite{You2019PositionawareGN} propose a method to add position information into GNN. 
Our methods are different from You \textit{et al.}. They model the \emph{absolute} position of a node in a full graph without any particular reference points; while our technique captures the \emph{relative} position of a node with respect to the query node. 
Finally, some work on graph generation \cite{Li2018LearningDG, Jin2018JunctionTV, You2018GraphCP} involves a module to add a new node into a partially generated graph, which shares the similar goal as our model. 
However, such graph generation model typically requires fully labeled training data to learn from. 
To the best of our knowledge, this is the first study on how to expand an existing directed acyclic graph (as we model a taxonomy as a DAG) using self-supervised learning. 

\section{Problem Formulation}\label{sec:problem}

In this section, we first define a taxonomy, then formulate our problem, and finally discuss the scope of our study.

\smallskip
\noindent \textbf{Taxonomy.} A taxonomy $\T = (\N, \E)$ is a directed acyclic graph where each node $n \in \N$ represents a concept (\ie, a word or a phrase) and each directed edge $\langle n_{p}, n_{c} \rangle \in \E$ indicates a relation expressing that concept $n_{p}$ is the most specific concept that is more general than concept $n_{c}$. 
In other words, we refer to $n_{p}$ as the \mquote{parent} of $n_{c}$ and $n_{c}$ as the \mquote{child} of $n_{p}$. 

\smallskip
\noindent \textbf{Problem Definition.} The input of the \textit{taxonomy expansion task} includes two parts: (1) an existing taxonomy $\T^{0} = (\N^{0}, \E^{0})$, and (2) a set of new concepts $\C$. 
This new concept set can be either manually specified by users or automatically extracted from text corpora. 
Our goal is to expand the existing taxonomy $\T^{0}$ into a 
larger taxonomy $\T = (\N^{0} \cup \C, \E^{0} \cup \R)$, where $\R$ is a set of newly discovered relations each including one new concept $c \in \C$. 

\begin{thm:eg}\label{ex:problem}
Figure~\ref{fig:task} shows an example of our problem. 
Given a field-of-study taxonomy $\T^{0}$ in the computer science domain and a set of new concepts $\C = \{\textit{``UDA''}, \textit{``Meta Learning''}, \dots\}$, we find each new concept's best position in $\T^{0}$ (\eg, ``UDA'' under ``Semi-supervised Learning'' as well as ``GPU'' under ``Integrated Circuit'') and expand $\T^{0}$ to include those new concepts.
\end{thm:eg}

\smallskip
\noindent \textbf{Simplified Problem.} A simplified version of the above problem is that we assume the input set of new concepts contains only one element (\ie, $|\C|=1$), and we aim to find one single parent node of this new concept (\ie, $|\R|=1$). We discuss the connection between these two problem settings at the end of Section~\ref{subsec:taxonomy_model}.

\smallskip
\noindent \textbf{Discussion.} 
In this work, we follow previous studies \cite{Jurgens2016SemEval2016T1, Vedula2018EnrichingTW, Aly2019EveryCS} and assume each concept in $\N^{0} \cup \C$ has an initial embedding vector learned from this concept's surface name, or if available, its definition sentences \cite{Schlichtkrull2016MSejrKuAS} and associated web pages \cite{Wang2014AHD}. 
We also note that our problem formulation assumes those relations in the existing taxonomy are not modified. 
We acknowledge that such modification is necessary in some cases, but it is much less frequent and requires high cautiousness from human curators. 
Therefore, we leave it out of the scope of automation in this study. 

\section{The TaxoExpan Framework}\label{sec:method}

In this section, we first introduce our taxonomy model and expansion goal.
Then, we elaborate how to represent a query concept and an insertion position (\ie, an anchor concept), based on which we present our query-concept matching model.
Finally, we discuss how to generate self-supervision data from the existing taxonomy and use them to train the \TaxoExpan framework. 

 \subsection{Taxonomy Model and Expansion Goal}\label{subsec:taxonomy_model}

A taxonomy $\T$ describes a hierarchical organization of concepts.
These concepts form the node set $\N$ in $\T$. 
Mathematically, we model each node $n \in \N$ as a categorical random variable and the entire taxonomy $\T$ as a Bayesian network.
We define the probability of a taxonomy $\T$ as the joint probability of node set $\N$ which can be further factorized into a set of conditional probabilities as follows:
\begin{displaymath}
\small
\P(\T | \Theta) = \P(\N | \T, \Theta) = \prod_{i=1}^{|\N|} \P(n_{i} | parent_{\T}(n_i), \Theta),
\end{displaymath}
where $\Theta$ is the set of model parameters and $parent_{\T}(n_{i})$ is the set of $n_i$'s parent node(s) in taxonomy $\T$. 

Given learned model parameters $\Theta$, an existing taxonomy $\T^{0} = (\N^{0}, \E^{0})$, and a set of new concepts $\C$, we can ideally find the best taxonomy $\T^{*}$ by solving the following optimization problem: 
\begin{displaymath}
\small
\T^{*} = \argmax_{\T} \P(\T | \Theta) = \argmax_{\T} \sum_{i=1}^{|\N^{0} \cup \C|} \log \P(n_{i} | parent_{\T}(n_i), \Theta).
\end{displaymath}

This \naive approach has two limitations. 
First, the search space of all possible taxonomies over the concept set $|\N^{0} \cup \C|$ is prohibitively large. 
Second, we cannot guarantee the structure of existing taxonomy $\T^{0}$ remains unchanged, which can be undesirable from the application point of view. 

We address the above limitations by restricting the search space of our output taxonomy to be the exact expansion of the existing taxonomy $\T^{0}$.
Specifically, we keep the parents of each existing taxonomy node $n \in \N^{0}$ unchanged and only try to find a \emph{single} parent node of each new concept in $\C$. 
As a result, we divide the above computationally intractable problem into the following set of $|\C|$ tractable optimization problems:
\begin{equation}\label{eq:opt}
\small
a_{i}^{*} = \argmax_{a_{i} \in \N^{0}} \log \P(n_{i} | a_{i}, \Theta), \quad \forall i \in \{1, 2, \dots, |\C| \},
\end{equation}
where $a_{i}$ is the parent node of a new concept $n_{i} \in \C$ and we refer to it as the \mquote{anchor concept}. 

\smallskip
\noindent \textbf{Discussion.}~
The above equation defines $|\C|$ \emph{independent} optimization problems and each problem aims to find one single parent of a new concept $n_{i}$. Therefore, we essentially reduce the more generic taxonomy expansion problem into $|C|$ independent simplified problems (c.f. Section~\ref{sec:problem}) and tackle it by inserting new concepts \emph{one-by-one} into the existing taxonomy.
As a result of the above reduction, possible interactions among new concepts are ignored and we leave it to the future work.
In the following sections, we continue to answer two keys questions: (1) how to model the conditional probability $\P(n_i | a_i, \Theta)$, and (2) how to learn model parameters $\Theta$. 


\begin{figure}[!t]
  \centering
  \centerline{\includegraphics[width=0.45\textwidth]{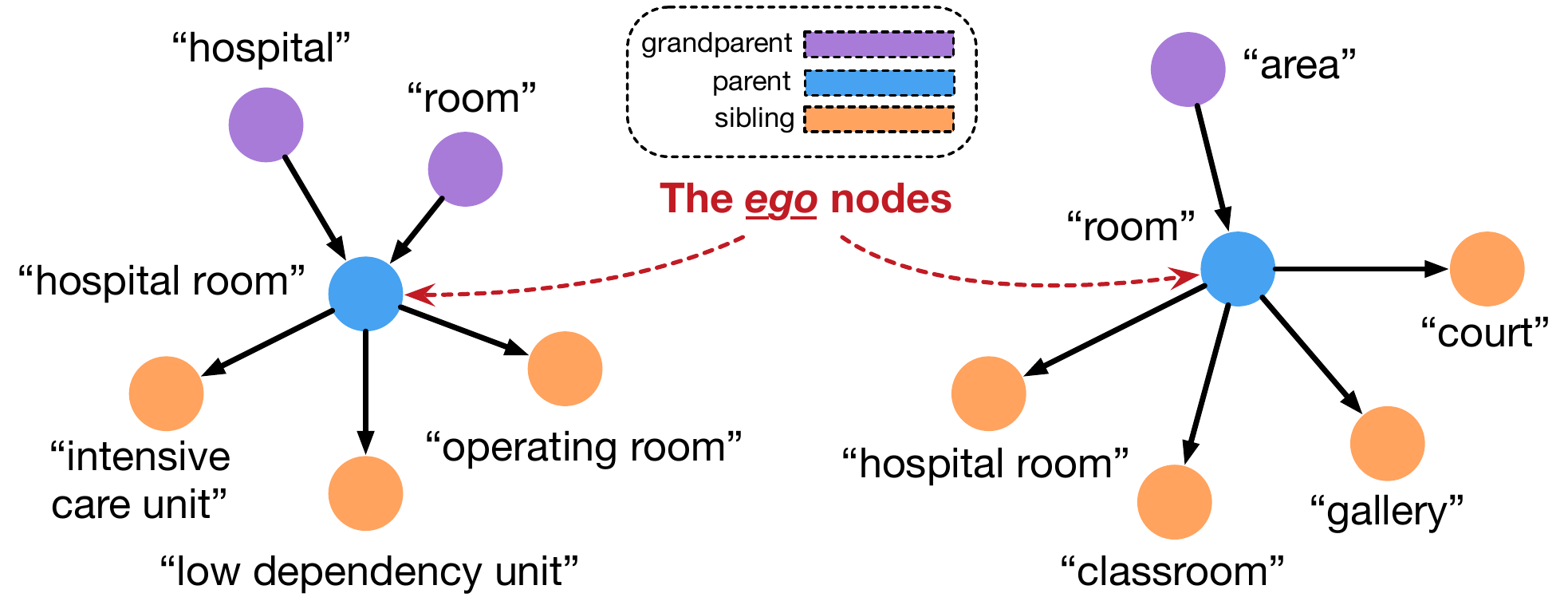}}
  \vspace{-0.2cm}
  \caption{Two egonets correspond to two anchor concepts.}
  \label{fig:egonet}
  \vspace{-0.2cm}
\end{figure}

\begin{figure*}[!t]
  \centering
  \centerline{\includegraphics[width=0.95\textwidth]{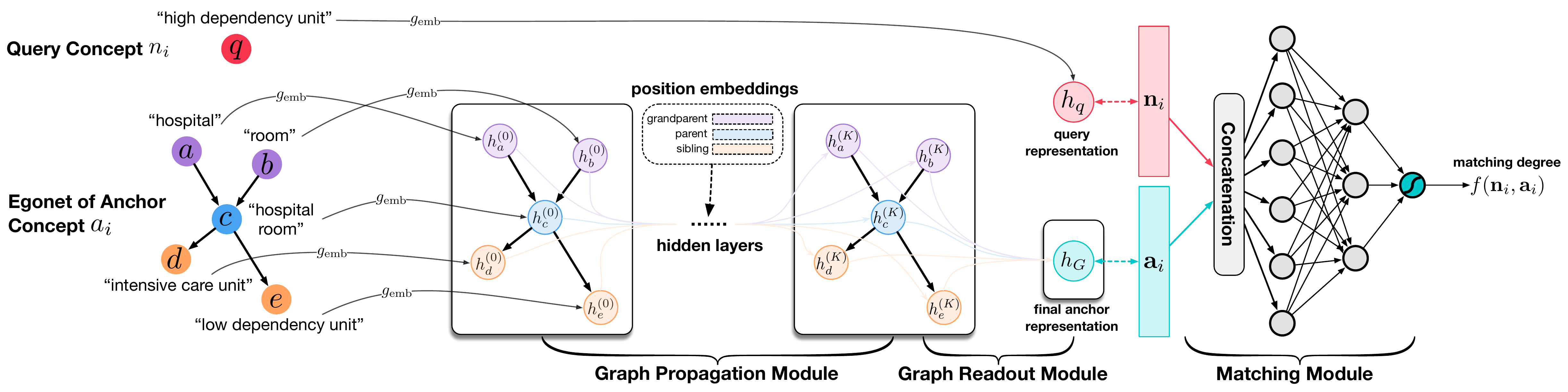}}
  \vspace{-0.2cm}
  \caption{Overview of \TaxoExpanBf framework. $g_{\text{emb}}$ is an embedding model that provides query concept's initial feature vector $h_{q}$ and the initial feature vector of each node in the egonet. The graph propagation module transforms initial feature vectors into better node representations based on which the graph readout module outputs the egonet embedding as the final anchor representation. Finally, a matching module inputs both query and anchor representations and outputs their matching score.}
  \label{fig:architecture}
  \vspace{-0.2cm}
\end{figure*}
\subsection{Modeling Query-Anchor Matching}\label{subsec:modeling_gnn}

We model the matching score between a query concept $n_i$ and an anchor concept $a_i$ by projecting them into a vector space and calculating matching scores using their vectorized representations. 
We show the entire model architecture of \TaxoExpan in Figure~\ref{fig:architecture}.

\subsubsection{Representing Query Concept}\label{subsubsec:query_representation}
\hfill

In this study, we assume each query concept has an \emph{initial feature vector} learned based on some text associated with this concept.
Such text can be as simple as the concept surface name, or in some prior studies \cite{Jurgens2016SemEval2016T1, Wang2014AHD}, the definition sentences and clicked web pages about the concept. 
We represent each query concept $n_{i}$ using its initial feature vector denoted as $\mathbf{n}_i$. 
We will discuss how to obtain such initial feature vectors using embedding learning methods in the experiment section. 


\subsubsection{Representing Anchor Concept}\label{subsubsec:query_representation}
\hfill

Each anchor concept corresponds to one node in the existing taxonomy $\T^{0}$ that could be the \mquote{parent} of a query concept. 
One \naive way to represent an anchor concept is to 
directly use its initial feature vector. 
A key limitation of this approach is that it captures only the \mquote{parent} node information and loses other surrounding nodes' signals, which could be crucial for determining whether the query concept should be put in this position. 
We illustrate this limitation below:
\begin{thm:eg}\label{ex:sibling_importance}
Suppose we are given a query concept \mquote{high dependency unit} to predict whether it should be under the \mquote{hospital room} node (\ie, an anchor concept) in an existing taxonomy. 
As these two concepts have dissimilar embeddings based on their surface names, we may believe this query concept shouldn't be placed underneath this anchor concept.
However, if we know that this anchor concept has two children nodes, \ie, \mquote{intensive care unit} and \mquote{low dependency unit}, that are closely related to the query concept, we are more likely to put the query concept under this anchor concept, correctly. 
\end{thm:eg}

The above example demonstrates the importance of capturing local structure information in the anchor concept representation. 
We model the anchor concept using its ego network. 
Specifically, we consider the anchor concept to be the \mquote{parent} node of a query concept. The ego network of the anchor concept consists of the \mquote{sibling} nodes and \mquote{grand parent} nodes of the query concept, as shown in Figure~\ref{fig:egonet}. 
We represent the anchor concept based on its ego network using a graph neural network.

\smallskip
\noindent \textbf{Graph Neural Network Architectures.}
Given an anchor concept $a_{i}$ with its corresponding ego network $G_{a_{i}}$ and its initial representation $a_i$, we use a graph neural network (GNN) to generate its final representation $\mathbf{a}_i$. 
This GNN contains two components: (1) a \emph{graph propagation} module that transforms and propagates node features over the graph structure to compute individual node embeddings in $G_{a_{i}}$, and (2) a \emph{graph readout} module that combines node embeddings into a vector representing the full ego network $G_{a_{i}}$. 
The final graph embedding encodes all local structure information centered around the anchor concept and we use it as the final anchor representation $\mathbf{a}_i$.

A graph propagation module uses a neighborhood aggregation strategy to iteratively update the representation of a node $u$ by aggregating representations of its neighbors $N(u)$ and itself. 
We denote $N(u) \cup \{u\}$ as $\wtd{N(u)}$.
After $K$ iterations, a node's representation captures the structural information within its $K$-hop neighborhood. 
Formally, we define a GNN with $K$-layers as follows:
\begin{equation}
\small
h_{u}^{(k)} = \text{AGG}^{(k)}\left(\{h_{v}^{(k-1)} | v \in \wtd{N(u)} \}\right), \quad k \in \{1,\dots, K \},
\end{equation}
where $h_{u}^{(k)}$ is node $u$'s feature in the $k$-th layer; $h_{u}^{(0)}$ is node $u$'s initial feature vector, and $\text{AGG}^{(k)}$ is an aggregation function in the $k$-th layer.
We instantiate $\text{AGG}^{(k)}$ using two popular architectures: Graph Convolutional Network (GCN) \cite{Kipf2017SemiSupervisedCW} and Graph Attention Network (GAT) \cite{Velickovic2018GraphAN}. 
GCN defines the $\text{AGG}$ function as follows:
\begin{equation}\label{eq:gcn_agg}
\small
\text{AGG}^{(k)}\left(\{h_{v}^{(k-1)} | v \in \wtd{N(u)} \}\right) = \rho\left( \sum_{v \in \wtd{N(u)}} \alpha_{uv}^{(k-1)} \mathbf{W}^{(k-1)} h_{v}^{(k-1)}  \right),
\end{equation}
where $\alpha_{uv}^{(k-1)} = 1/$ $\small{\sqrt{|\wtd{N(u)}||\wtd{N(v)}|}}$ is a normalization constant (same for all layers); $\rho$ is a non-linear function (\eg, ReLU), and $\mathbf{W}^{(k-1)}$ is the learnable weight matrix. 
If we interpret $\alpha_{uv}^{(k-1)}$ as the \emph{importance} of node $v$'s feature to node $u$, GCN calculates it using only the graph structure without leveraging the node features. 
GAT addresses this limitation by defining $\alpha_{uv}^{(k-1)}$ as follows:
\begin{equation}\label{eq:gat_alpha}
\small
\alpha_{uv}^{(k-1)} = \frac{\exp\left(\gamma\left(\mathbf{z}^{(k-1)}\dot [\mathbf{W}^{(k-1)} h_{u}^{(k-1)} \| \mathbf{W}^{(k-1)} h_{v}^{(k-1)}]\right)\right)}{\sum_{v' \in \wtd{N(u)}} \exp\left(\gamma\left(\mathbf{z}^{(k-1)}\dot [\mathbf{W}^{(k-1)} h_{u}^{(k-1)} \| \mathbf{W}^{(k-1)} h_{v'}^{(k-1)}]\right)\right)},
\end{equation}
where both $\mathbf{z^{(k-1)}}$ and $\mathbf{W}^{(k-1)}$ are learnable parameters; $\gamma(\cdot)$ is another non-linear function (\eg, LeakyReLU), and ``$\|$'' represents the concatenation operation. 
Plugging the above $\alpha_{uv}^{(k-1)}$ into Eq. (\ref{eq:gcn_agg}) we obtain the aggregation function in a \emph{single-head} GAT. 
Finally, We execute $M$ independent transformations of Eq. (\ref{eq:gcn_agg}) and concatenate their output features to compose the final output embedding of node $u$. 
This defines the aggregation function in a \emph{multi-head} GAT (with $M$ heads) as follows:
\begin{equation}\label{eq:gat_agg}
\small
\hspace{-0.2cm} \text{AGG}^{(k)}\left(\{h_{v}^{(k-1)} | v \in \wtd{N(u)} \}\right) = \hspace{-0.1cm} \concat_{m=1}^{M} \rho \left( \sum_{v \in \wtd{N(u)}} \alpha_{uv}^{(k-1)} \mathbf{W}^{(k-1)}_{m} h_{v}^{(k-1)}  \right),
\end{equation}
where $\mathbf{W}^{(k-1)}_{m}$ is the $m$-th weight matrix in the $m$-th attention head. 

After obtaining each node's final representation $h_{u}^{(K)}$, we generate the ego network's representation $h_{G}$ using a graph readout module as follows:
\begin{equation}\label{eq:gnn_readout}
\small
h_{G} = \text{READOUT}(\{h_{u}^{(K)} | u \in G\}),
\end{equation}
where $\text{READOUT}$ is a permutation invariant function \cite{Zaheer2017DeepS} such as element-wise mean or sum.

\smallskip
\noindent \textbf{Position-enhanced Graph Neural Networks.}
One key limitation of the above GNN model is that they fail to capture each node's position information relative to the query concept.  
Take Figure~\ref{fig:egonet} as an example, the \mquote{hospital room} node in the left ego network 
is the anchor node itself
while in the right ego network it 
is the child of the anchor node.
Such position information will influence how node feature propagates within the ego network and how the final graph embedding is aggregated. 

An important innovation in \TaxoExpan is the design of position-enhanced graph neural networks. 
The key idea is to learn a set of ``position embeddings'' and enrich each node feature with its corresponding position embedding. 
We denote node $u$'s position as $p_u$ and its position embedding at $k$-th layer as $\mathbf{p}_u^{(k)}$. 
We replace each node feature $h_{u}^{(k-1)}$ with its position-enhanced version $h_{u}^{(k-1)} \| \mathbf{p}_u^{(k-1)}$ in Eqs. (\ref{eq:gcn_agg}-\ref{eq:gat_agg}) and adjust the dimensionality of $\mathbf{W}^{(k-1)}$ accordingly. 
Such position embeddings help us to learn better node representations from two aspects. 
First, we can capture more neighborhood information. 
Take $\mathbf{W}^{(k-1)} h_{v}^{(k-1)}$ in the right hand side of Eq. (\ref{eq:gcn_agg}) as an example, we enhance it to the following:
\begin{displaymath}\label{eq:pgnn_agg}
\small
\left[\mathbf{W}^{(k-1)} \| \mathbf{O}^{(k-1)}\right]\left[h_{v}^{(k-1)} \| \mathbf{p}_{v}^{(k-1)}\right]  = \mathbf{W}^{(k-1)}h_{v}^{(k-1)} + \mathbf{O}^{(k-1)}\mathbf{p}_{v}^{(k-1)}, 
\end{displaymath}
where $\mathbf{O}^{(k-1)}$ is another weight matrix used to transform position embeddings. 
The above equation shows that a node's new representation is jointly determined by its neighborhoods' contents (\ie, $h_{v}^{(k-1)}$) and relative positions in the ego network (\ie, $\mathbf{p}_{v}^{(k-1)}$). 
Second, for GAT architecture, we can better model neighbor importance as the term $\alpha_{uv}^{(k-1)}$ in Eq. (\ref{eq:gcn_agg}) currently depends on both $\mathbf{p}_{u}^{(k-1)}$ and  $\mathbf{p}_{v}^{(k-1)}$. 

Furthermore, we propose two schemes to inject position information in the graph readout module. 
The first one, called weighted mean readout (WMR), is defined as follows:
\begin{equation}\label{eq:readout_wmr}
\small
\text{READOUT}(\{h_{u}^{(K)} | u \in G\}) = \sum_{u \in G} \frac{\log(1+\exp(\alpha_{p_u}))}{\sum_{u' \in G} \log(1+\exp(\alpha_{p_u'}))} h_{u}^{(K)}, 
\end{equation}
where $\alpha_{p_{u}}$ is the parameter indicating the importance of position $p_{u}$.
The second scheme is called concatenation readout (CR) which combines the average embeddings of nodes with the same position as follows:
\begin{equation}\label{eq:readout_car}
\small
\text{READOUT}(\{h_{u}^{(K)} | u \in G\}) = \concat_{p \in \mathcal{P}} \frac{\mathcal{I}(p_{u} = p) h_{u}^{(K)}}{\sum_{u' \in G}\mathcal{I}(p_{u'} = p)},
\end{equation}
where $\mathcal{P}$ is the set of all positions we are modeling and $\mathcal{I}(\cdot)$ is an indicator function which returns 1 if its internal statement is true and returns 0 otherwise. 

\subsubsection{Matching Query Concept and Anchor Concept}\label{subsubsec:query_anchor_matching}
\hfill

Based on the learned query concept representation $\mathbf{n}_i \in \mathbb{R}^{D_1}$ and anchor concept representation $\mathbf{a}_i \in \mathbb{R}^{D_2}$, we calculate their match score using a matching module $f(\cdot): \mathbb{R}^{D_2} \times \mathbb{R}^{D_1} \rightarrow \mathbb{R}$.
We study two architectures. 
The first one is a multi-layer perceptron with one hidden layer, defined as follows:
\begin{equation}\label{eq:match_mlp}
\small
f^{\text{MLP}}(\mathbf{a}_i, \mathbf{n}_i) = \sigma\left(\mathbf{W}_2 \gamma (\mathbf{W}_1 ( \mathbf{a}_i \| \mathbf{n}_i ) + \mathbf{B}_1)+ \mathbf{B}_2 \right),
\end{equation}
where $\{\mathbf{W}_1, \mathbf{B}_1, \mathbf{W}_2, \mathbf{B}_2 \}$ are parameters; $\sigma(\cdot)$ is the sigmoid function, and $\gamma(\cdot)$ is the LeakyReLU activation function.
The second architecture is a log-bilinear model defined as follows:
\begin{equation}\label{eq:match_lb}
\small
f^{\text{LBM}}(\mathbf{a}_i, \mathbf{n}_i) = \exp\left(\mathbf{a}_i^{T} \mathbf{W} \mathbf{n}_i \right),
\end{equation}
where $\mathbf{W}$ is a learnable interaction matrix. 
We choose these MLP and LBM as they are representative architecures in linear and bilinear interaction models, respectively.

\subsection{Model Learning and Inference}\label{subsec:learning}

\begin{figure}[!t]
  \centering
  \centerline{\includegraphics[width=0.49\textwidth]{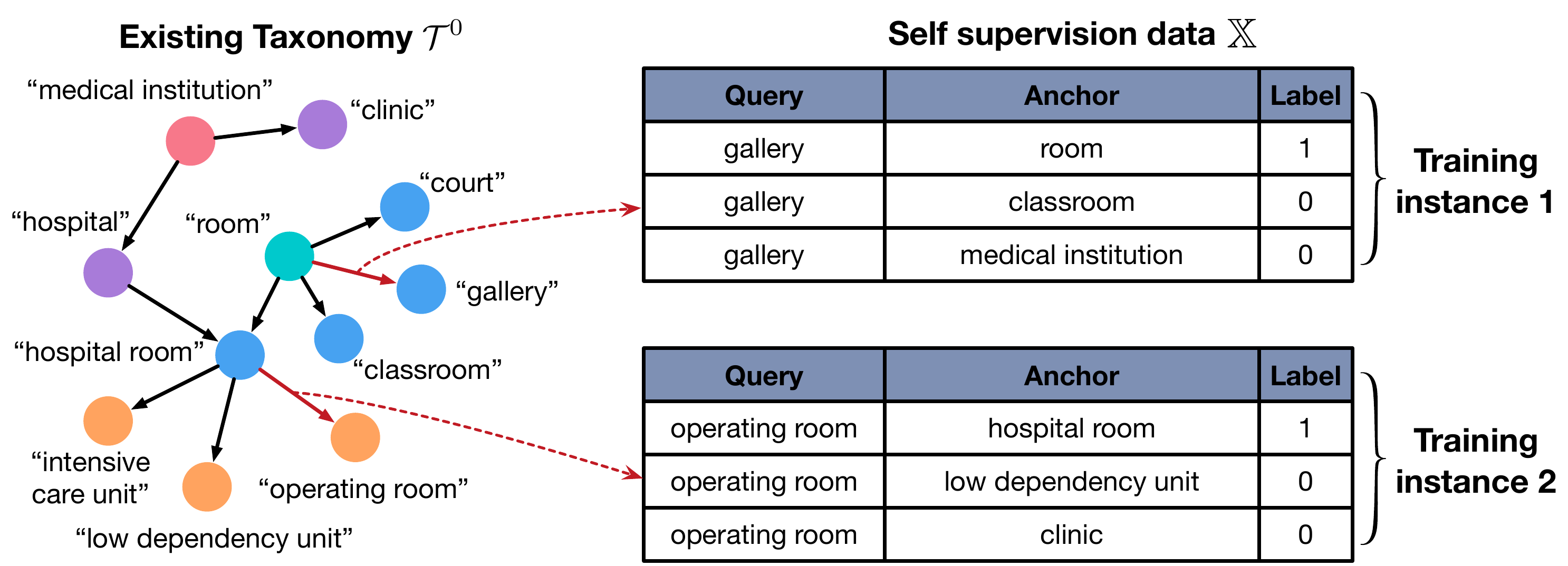}}
  \vspace{-0.2cm}
  \caption{Self-supervision generation.}
  \label{fig:supervision}
  \vspace{-0.2cm}
\end{figure}

The above section discusses how to model query-anchor matching using a parameterized function $f(\cdot | \Theta)$. 
In this section, we first introduce how we learn those parameters $\Theta$ using self-supervision from the existing taxonomy.
Then, we establish the connection between the matching score with the conditional probability $\P(n_i | a_i)$, and discuss how to conduct model inference. 

\smallskip
\noindent \textbf{Self-supervision Generation.}
Figure~\ref{fig:supervision} shows the generation process of self supervision data. 
Given one edge $\pair{n_p, n_c}$ in the existing taxonomy $\T^{0} = (\N^{0}, \E^{0})$, we first construct a positive $\pair{\text{anchor, query}}$ pair by using child node $n_c$ as the \mquote{query} and parent node $n_p$ as the \mquote{anchor}.
Then, we construct $N$ negative pairs by fixing the query node $n_{c}$ and randomly selecting $N$ nodes $\{n_{r}^{l}|_{l=1}^{N}\} \subset \N^{0}$ that are neither parents nor descendants of $n_{c}$. 
These $N+1$ pairs (one positive and $N$ negatives) collectively consist of one training instance $\mathbf{X} = \{\pair{n_p, n_{c}}, \pair{n_{r}^{1}, n_c}, \dots, \pair{n_{r}^{N}, n_c} \}$. 
By repeating the above process for each edge in $\T^{0}$, we obtain the full self-supervision dataset $\mathbb{X} = \{\mathbf{X_1}, \dots, \mathbf{X_{|\E^{0}|}} \}$.
Notice that a node with $C$ parents in $\T^{0}$ will derive $C$ training instances in $\mathbb{X}$.

\smallskip
\noindent \textbf{Model Training.}
We learn our model on $\mathbb{X}$ using the InfoNCE loss~\cite{Oord2018RepresentationLW} as follows:
\begin{equation}\label{eq:loss}
\small
\mathcal{L}(\Theta) = -\frac{1}{|\mathbb{X}|} \sum_{\mathbf{X}_i \in \mathbb{X}} \left[ \log\frac{f(n_p, n_c)}{\sum_{\pair{n_j, n_c} \in \mathbf{X}_i} f(n_j, n_c)}     \right],
\end{equation}
where the subscript $j \in [1, 2, \dots, N+1]$. If $j=1$, $\pair{n_j, n_c}$ is a positive pair, otherwise, $\pair{n_j, n_c}$ is a negative pair. 
The above loss is the cross entropy of classifying the positive pair $\pair{n_p, n_c}$ correctly, with $\frac{f(n_p, n_c)}{\sum_{\pair{n_j, n_c} \in \mathbf{X}_i} f(n_j, n_c)}$ as the model prediction.
Optimizing this loss results in $f(a_i, n_i)$ estimating the following probability density (up to a multiplicative constant):
\begin{equation}\label{eq:density}
\small
f(a_i, n_i) \propto \frac{\P(a_i | n_i)}{\P(a_i)}.
\end{equation}
We prove the above result in Appendix and summarize our self-learning procedure in Algorithm~\ref{algo:self_train}. 
We establish the connection between matching score $f(a_i, n_i)$ with the probability $\P(n_i | a_i)$ in Eq.~\ref{eq:opt} as follows:
\begin{equation}\label{eq:bayes}
\small
\P(n_i  | a_i) = \frac{\P(a_i | n_i)}{\P(a_i)} \cdot \P(n_i) \propto f(a_i, n_i) \cdot \P(n_i). 
\end{equation}
We elaborate the implication of this equation below.

\smallskip
\noindent \textbf{Model Inference.}
At the inference stage, we are given a new query concept $n_i$ and apply the learned model $f(\cdot|\Theta)$ to predict its parent node in the existing taxonomy $\T^{0}$. 
Mathematically, we aim to find the anchor position $a_{i}$ that maximizes $\P(n_i|a_i)$, which is equivalent to maximizing $f(a_i, n_i)$ because of Eq. (\ref{eq:bayes}) and the fact that $P(n_i)$ is the same across all positions. 
Therefore, we rank all candidate positions $a_i$ based on their matching scores with $n_i$ and select the top ranked one as the predicted parent node of this query concept. 
Although we currently select only the top one as query's single parent, we can also choose top-$k$ ones as query's parents, if needed.

\smallskip
\noindent \textbf{Summary.}
Given an existing taxonomy and a set of new concepts, our \TaxoExpan first generates a set of self-supervision data and learns its internal model parameters using Algorithm~\ref{algo:self_train}.
For each new concept, we run the inference procedure and find its best parent node in the existing taxonomy. 
Finally, we place these new concepts underneath their predicted parents one at a time, and output the expanded taxonomy. 

\smallskip
\noindent \textbf{Computational Complexity Analysis.}
At the training stage, our model uses $|\E^{(0)}|$ training instances every epoch and thus scales linearly to the number of edges in the existing taxonomy. 
At the inference stage, for each query concept, we calculate $|\N^{(0)}|$ matching scores, one for every existing node in $\T^{0}$. 
Although such $O(|\N^{(0)}|)$ cost per query is expensive, we can significantly reduce it using two strategies. 
First, most computation efforts of \TaxoExpan are matrix multiplications and thus we use GPU for acceleration.
Second, as the graph propagation and graph readout modules are query-independent (c.f. Fig. \ref{fig:supervision}), we pre-compute all anchor representations and cache them. 
When a set of queries are given, we only run the matching module. 
In practice, it takes less than 30 seconds to calculate all matching scores between 2,450 queries with over 24,000 anchor positions on a single K80 GPU.

 \begin{algorithm}[!t]
   \small
   \caption{Self-supervised learning of \TaxoExpan}
   \label{algo:self_train}
   \KwIn{
     A taxonomy $\T^{0}$; negative size $N$, batch size $B$; model $f(\cdot|\Theta)$.
   }
   \KwOut{Learned model parameters $\Theta$.}
   Randomly initialize $\Theta$\;
   \While{$\mathcal{L}(\Theta)$ in Eq. (\ref{eq:loss}) not converge} {
   	Enumerate edges in $\T^{0}$ and sample $B$ edges without replacement\;
	$\mathbb{X} = \{\}$ \# current batch of training instances\;
	\For{each sampled edge $\pair{n_p, n_c}$} {
		Generate $N$ negative pairs $\{\pair{n_{r}^{l}, n_c}|_{l=1}^{N}\}$\;
		$\mathbb{X} \gets \mathbb{X} \cup \{\pair{n_p, n_c}, \pair{n_{r}^{1}, n_c}, \dots, \pair{n_{r}^{N}, n_c}\}$\;
	}
	Update $\Theta$ based on $\mathbb{X}$.
   }
   Return $\Theta$\;
 \end{algorithm}
 
\section{Experiments}\label{sec:exp}

In this section, we study the performance of \TaxoExpan on three large-scale real-world taxonomies.

%
%
%

\subsection{Expanding MAG Field-of-Study Taxonomy}
\subsubsection{Datasets}
We evaluate \TaxoExpan on the public Field-of-Study (FoS) Taxonomy\footnote{\scriptsize \url{https://docs.microsoft.com/en-us/academic-services/graph/reference-data-schema}} in Microsoft Academic Graph (MAG)~\cite{Sinha2015AnOO}. 
This FoS taxonomy contains over 660 thousand scientific concepts and more than 700 thousand taxonomic relations.
Although being constructed semi-automatically, this taxonomy is of high quality, as shown in the previous study \cite{Shen2018AWS}. Thus we treat each concept's original parent nodes as its correct anchor positions. 
We remove all concepts that have no relation in the original FoS taxonomy and then randomly mask 20\% of leaf concepts (along with their relations) for validation and testing\footnote{\scriptsize Here we mask only leaves because if we remove intermediate nodes, we have to remove their descendants from the candidate parent pool, which causes different masked nodes (as testing query concepts) having different candidate pools.}.
The remaining FoS taxonomy is then treated as the input existing taxonomy. 
We refer to this dataset as \textbf{MAG-Full}. 
Based on MAG-Full, we construct another dataset focusing on the computer science domain. 
Specifically, we first select a subgraph consisting of all descendants of ``computer science'' node and then mask 10\% of leaf concepts in this subgraph for validation and another 10\% of leaf nodes for testing. 
We name this dataset as \textbf{MAG-CS}. 

\begin{table}[!t]
    \centering
    \caption{Dataset Statistics. $|\N|$ and $|\E|$ are the number of nodes and edges in the existing taxonomy. $|\D|$ indicates the taxonomy depth and $|\C|$ is the number of new concepts.}
    \label{tbl:dataset}
    \vspace{-0.2cm}
    \scalebox{0.9}{
        \begin{tabular}{ccccc}
            \toprule
            \textbf{Dataset}    & $|\N|$ &  $|\E|$ & $|\D|$ & $|\C|$ \\
            \midrule
            \textbf{MAG-CS}      & 24,754     &  42,329   & 6    & 2,450 \\
            \textbf{MAG-Full}    & 355,808    &   638,674   &   6    & 37,804 \\
            \textbf{SemEval}   & 95,882      & 89,089      & 20      & 600  \\
    	\bottomrule
        \end{tabular}
    }
    \vspace{-0.2cm}
\end{table}

\begin{table*}[!t]
	\centering
	\caption{Overall results on MAG-CS and MAG-Full datasets. We run all methods three times and report the averaged result with standard deviation. Note that smaller MR indicates better model performance. For all other metrics, larger values indicate better performance. We highlight \underline{the best two models} in terms of the average performance under each metric.}
	\label{tbl:mag_overall_results}
	\vspace{-0.2cm} 
	\scalebox{0.78}{
        \begin{tabular}{c|cccc|cccc}
        		\toprule
             	\multirow{2}{*}{\textbf{Method}} & \multicolumn{4}{c}{\textbf{MAG-CS}} & \multicolumn{4}{c}{\textbf{MAG-Full}} \\
		\cmidrule{2-9}
		& MR & Hit@1 & Hit@3 & MRR   & MR & Hit@1 & Hit@3 & MRR   \\
		\midrule
		Closest-Parent      & 1327.16 ($\pm$0.000)  & 0.0531 ($\pm$0.000)   & 0.0986 ($\pm$0.000)   & 0.2691 ($\pm$0.000)	& 14355.5 ($\pm$0.000) & 0.0360 ($\pm$0.000) & 0.0728 ($\pm$0.000) & 0.1897 ($\pm$0.000)\\
		Closest-Neighbor  & 382.07 ($\pm$0.000)    & 0.1085 ($\pm$0.000)   & 0.2000 ($\pm$0.000)   & 0.3987 ($\pm$0.000)	& 4160.8 ($\pm$0.000) & 0.0221 ($\pm$0.000) & 0.0419 ($\pm$0.000) &0.1405 ($\pm$0.000)\\
		\midrule
		dist-XGBoost       & 136.86 ($\pm$1.832)    & 0.1903 ($\pm$0.010)   & 0.3483 ($\pm$0.014)   & 0.6618 ($\pm$0.003) 	& \textbf{426.70 ($\pm$8.047)} & \textbf{0.1498 ($\pm$0.076)} & \textbf{0.3046 ($\pm$0.009)} & 0.5621 ($\pm$0.002)\\
		ParentMLP          & \textbf{114.79 ($\pm$12.25)}    & 0.0729 ($\pm$0.088)   & 0.2656 ($\pm$0.037)   & 0.6454 ($\pm$0.009)		& 457.14 ($\pm$39.81) & 0.098 ($\pm$0.094) &  0.1928 ($\pm$0.086) & 0.4950 ($\pm$0.012)\\
		DeepSetMLP       & 115.26 ($\pm$9.159)    & \textbf{0.1988 ($\pm$0.005)}   & \textbf{0.3581 ($\pm$0.016)}   & \textbf{0.6653 ($\pm$0.015)} 		& 444.83 ($\pm$27.59) & 0.1461 ($\pm$0.005) & 0.2971 ($\pm$0.064) & \textbf{0.6392 ($\pm$0.017)}\\
		\midrule
 		\TaxoExpanBf      & \textbf{80.33 ($\pm$5.470)}    & \textbf{0.2121 ($\pm$0.010)}   & \textbf{0.3823 ($\pm$0.012)}   & \textbf{0.6929 ($\pm$0.003)} 		& \textbf{341.31 ($\pm$33.62)} & \textbf{0.1523 ($\pm$0.009)} & \textbf{0.3087 ($\pm$0.010)} & \textbf{ 0.6453 ($\pm$0.035)}\\
		\bottomrule
         \end{tabular}
 	}
	\vspace{-0.2cm}
\end{table*} 

To obtain the initial feature vector, we first construct a corpus that consists of all paper abstracts mentioning at least one concept in the original MAG dataset.
Then, we use~``$\underline{\hspace{0.08in}}$'' to concatenate all tokens in one concept  (\eg, ``machine learning''~$\rightarrow$ ``machine\_learning'') and learn 250-dimension word embeddings using skipgram model in word2vec\footnote{\scriptsize We also test CBOW model, fastText \cite{bojanowski2016enriching} and BERT embedding \cite{Devlin2018BERTPO} (averaged across all concept mentions), and empirically we find skipgram model in word2vec works best on this dataset.} \cite{Mikolov2013DistributedRO}. 
Finally, we use these learned embeddings as the initial feature vector. 
Table~\ref{tbl:dataset} lists the statistics of these two datasets. 
All datasets and our model implementations are available at: \textcolor{blue}{\url{https://github.com/mickeystroller/TaxoExpan}}.

\vspace{-0.1cm}
\subsubsection{Evaluation Metrics}
As our model returns a rank list of all candidate parents for each input query concept, we evaluate its performance using the following three ranking-based metrics.
\begin{itemize}[leftmargin=*]
\item \textbf{Mean Rank (MR)} measures the average rank position of a query concept's true parent among all candidates. For queries with multiple parents, we first calculate the rank position of each individual parent and then take the average of all rank positions. Smaller MR value indicates better model performance. 
\item \textbf{Hit@}$k$ is the number of query concepts whose parent is ranked in the top $k$ positions, divided by the total number of queries. 
\item \textbf{Mean Reciprocal Rank (MRR)} calculates the reciprocal rank of a query concept's true parent. We follow \cite{Ying2018GraphCN} and use a scaled version of MRR in the below equation:
\begin{displaymath}
\small
\text{MRR} = \frac{1}{|\C|}\sum_{c \in \C} \frac{1}{|parent(c)|}\sum_{i \in parent(c)} \frac{1}{\lceil R_{i, c} / 10 \rceil},
\end{displaymath}
where $parent(c)$ represents the parent node set of the query concept $c$, and $R_{i, c}$ is the rank position of query concept $c$'s true parent $i$. 
We scale the original MRR by a factor 10 in order to amplify the performance gap between different methods. 
\end{itemize}

\vspace{-0.2cm}
\subsubsection{Compared Methods}\label{subsec:compare_methods}
We compare the following methods:
\begin{enumerate}[leftmargin=*]
\item \textbf{Closest-Parent}: A rule-based method which first scores each candidate position in the existing taxonomy based on its cosine distance to the query concept between their initial embedding, and then ranks all positions using this score. 
The position with the smallest distance is chosen to be query concept's parent. 
\item \textbf{Closest-Neighbor}: Another rule-based method that scores each position based on its distance to the query concept plus the average distance between its children nodes and the query. 
\item \textbf{dist-XGBoost}: A self-supervised boosting method that works directly on 39 manually-designed features generated using initial node embeddings without any embedding transformation.
We input these features into XGBoost \cite{Chen2016XGBoostAS}, a tree-based boosting model, to predict the matching score between a query concept and a candidate position. 
\item \textbf{ParentMLP}: A self-supervised method that first concatenates the query concept embedding with the candidate position embedding and then feeds them into a Multi-Layer Perceptron (MLP) for prediction. 
\item \textbf{DeepSetMLP}: Another self-supervised method that extends ParentMLP by adding information of candidate position's children nodes. Specifically, we first use DeepSet architecture \cite{Zaheer2017DeepS} to generate the representation of the children node set and then concatenate it with query \& candidate position representations before the final MLP module. 
\item \TaxoExpanBf: Our proposed framework using position-enhanced GAT (PGAT) as graph propagation module, weighted mean readout (WMR) for graph readout, and log-bilinear model (LBM) for query-anchor matching. We learn this model using our proposed InfoNCE loss. 
\end{enumerate}

\vspace{-0.2cm}
\subsubsection{Implementation Details and Parameter Settings}\label{subsubsec:implementation_details}
For a fair comparison, we use the same 250-dimension embeddings across all compared methods. 
We use Google's original word2vec implementation\footnote{\scriptsize \url{https://github.com/tmikolov/word2vec}} for learning embeddings and employ gensim\footnote{\scriptsize \url{https://github.com/RaRe-Technologies/gensim}} to load trained embeddings for calculating term distances in Closest-Parent, Closest-Neighbor, and dist-XGBoost methods. 
For the other three methods, we implement them using PyTorch and DGL framework\footnote{\scriptsize \url{https://github.com/dmlc/dgl}}. 
We tune hyper-parameters in all self-supervised methods on the masked validation set. 
For \TaxoExpan, we use a two-layer position-enhanced GAT where the first layer has four attention heads (of size 250) and the second layer has one attention head (of size 500). 
For both layers, we use 50-dimension position embeddings and apply dropout with rate 0.1 on the input feature vectors. 
We use Adam optimizer with initial learning rate 0.001 and ReduceLROnPlateau scheduler\footnote{\scriptsize \url{https://pytorch.org/docs/stable/optim.html\#torch.optim.lr\_scheduler.ReduceLROnPlateau}} with three patience epochs. 
We discuss the influence of these hyper-parameters in the next subsection. 

\begin{table}[!t]
	\centering
	\caption{Ablation analysis of model architectures on MAG-CS dataset. We assign an index to each model variant (shown in the first column). All models are run three times with their averaged scores reported.}
	\label{tbl:ablation_analysis_architecture}
	\vspace{-0.2cm} 
	\scalebox{0.78}{
        \begin{tabular}{c|ccc|cccc}
        		\toprule
		\multirow{2}{*}{Ind} & Graph 		& Graph 		& \multirow{2}{*}{Matching} & \multirow{2}{*}{MR} & \multirow{2}{*}{Hit@1} & \multirow{2}{*}{Hit@3} & \multirow{2}{*}{MRR} \\
		  & Propagate	& Readout 	& 						      &                                            &						    &						     & \\
		\midrule
		1 & GCN 	& Mean	& MLP & 167.82 & 0.1581   & 0.2964 & 0.6002 \\
		2 & GAT  	& Mean	& MLP & 131.46 & 0.1584   & 0.3192 & 0.6409 \\
		3 & PGCN 	& Mean	& MLP & 148.54 & 0.1809   & 0.3015 & 0.6255 \\
		\textbf{4} & \textbf{PGAT} 	& \textbf{Mean}	& \textbf{MLP} & \textbf{100.80} & \textbf{0.1896}   & \textbf{0.3304} & \textbf{0.6525} \\
		\midrule
		5 & PGCN 	& WMR & MLP & 144.81 & 0.1798   & 0.3014 & 0.6309 \\
		6 & PGCN 	& CR	& MLP & 135.89 & 0.1902   & 0.3118 & 0.6348 \\
		\textbf{7} & \textbf{PGAT} 	& \textbf{WMR} & \textbf{MLP} & \textbf{92.62} & \textbf{0.1945}   & \textbf{0.3584} & \textbf{0.6619} \\
		8 & PGAT 	& CR	& MLP & 95.84 & 0.1897   & 0.3512 & 0.6596 \\
		\midrule
		9 & PGCN 	& WMR & LBM & 139.41 & 0.1829   & 0.3370 & 0.6642 \\
		10 & PGCN 	& CR	& LBM & 130.12 & 0.1934   & 0.3462 & 0.6776 \\
		\textbf{11} & \textbf{PGAT} 	& \textbf{WMR} & \textbf{LBM} & \textbf{80.33} & \textbf{0.2121}   & \textbf{0.3823} & \textbf{0.6929} \\
		12 & PGAT 	& CR	& LBM & 84.40 & 0.2089   & 0.3813 & 0.6894 \\
		\bottomrule
         \end{tabular}
 	}
	\vspace{-0.2cm}
\end{table} 

\vspace{-0.1cm}
\subsubsection{Experimental Results.}  
We present the experimental results in the following aspects. 

\smallskip
\noindent \textbf{1. $~~$Overall Performance.}
Table~\ref{tbl:mag_overall_results} presents the results of all compared methods. 
First, we find that Closest-Neighbor method clearly outperforms Closest-Parent method and DeepSetMLP is much better than ParentMLP. This demonstrates the effectiveness of modeling local structure information. 
Second, we compare dist-XGBoost method with Closest-Neighbor and show that self-supervision indeed helps us to learn an effective way to combine various neighbor distance information. 
All four self-supervised methods outperform rule-based methods.
Finally, our proposed \TaxoExpan has the overall best performance across all the metrics and defeats the second best method by a large margin.

\smallskip
\noindent \textbf{2. $~~$Ablation Analysis of Model Architectures.}
\TaxoExpan contains three key components: a graph propagation module, a graph readout module, and a matching model. 
Here, we study how different choices of these components affect the performance of \TaxoExpan.
Table~\ref{tbl:ablation_analysis_architecture} lists the results and the first column contains the index of each model invariant. 

First, we analyze graph propagation module by using simple average scheme for graph readout and MLP for matching. 
By comparing model 1 to model 3 and model 2 to model 4, we can see that graph attention architecture (GAT) is better than graph convolution architecture (GCN).
Furthermore, the position-enhanced variants clearly outperform their non-position counterparts (model 3 versus model 1 and model 4 versus model 2). This illustrates the efficacy of the position embeddings in the graph propagation module.

Second, we study graph readout module by fixing the graph propagation module to be the best two variants among models 1-4.
We can see both model 5 \& 6 outperform model 3 and model 7 \& 8 outperform model 4. This signifies that the position information also helps in the graph readout module.
However, the best strategy of incorporating position information depends on the graph propagation module. 
The concatenation readout scheme works better for PGCN while the weighted mean readout is better for PGAT. 
One possible explanation is that the concatenation readout leads to more parameters in matching model and as PGAT itself has more parameters than PGCN, further introducing more parameters in PGAT may cause the model to be overfitted. 

Finally, we examine the effectiveness of different matching models.
We replace the MLP in models 5-8 with LBM to create model variants 9-12. 
We can clearly see that LBM works better than MLP. It could be that LBM better captures the interaction between the query representation and the final anchor representation.

\begin{figure}[!t]
  \centering
  \centerline{\includegraphics[width=0.48\textwidth]{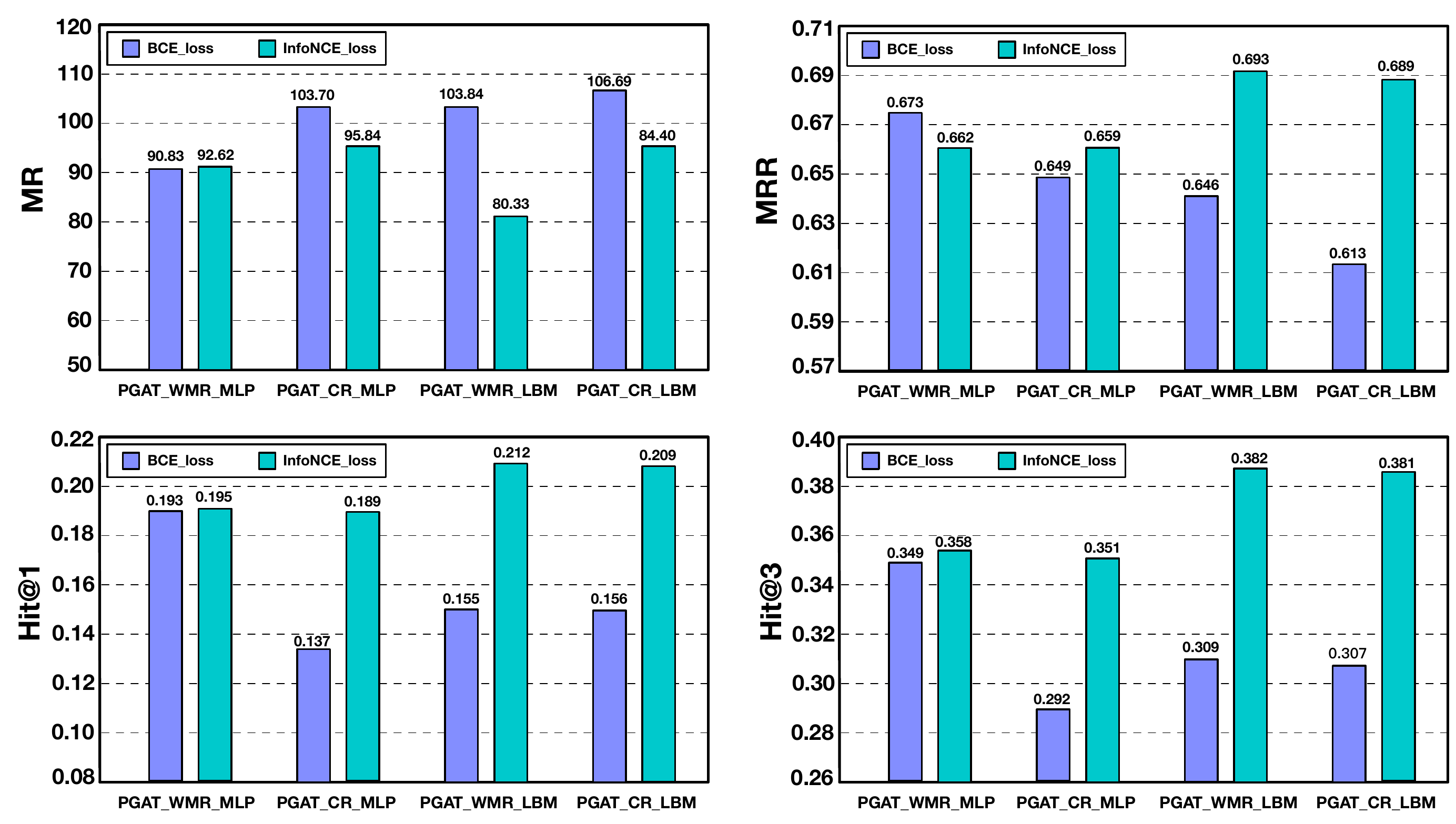}}
  \vspace{-0.2cm}
  \caption{Ablation analysis of training schemes on MAG-CS dataset. We compare models trained using Binary Cross Entropy (BCE) loss with those trained using InfoNCE loss.}
  \label{fig:ablation_analysis_loss}
  \vspace{-0.3cm}
\end{figure}

\smallskip
\noindent \textbf{3. $~~$Ablation Analysis of Training Schemes.}
In this subsection, we evaluate the effectiveness of our proposed training scheme. 
In this study, we first group a set of positive and negative $\pair{query, anchor}$ pairs into \emph{one single} training instance (c.f. Sect.~\ref{subsec:learning}) and learn the model using InfoNCE loss (c.f. Eq. (\ref{eq:loss})).
An alternative is to treat these pairs as different instances and train the model using standard binary cross entropy (BCE) loss.
Under this training scheme, we formulate our problem as a binary classification task. 
We compare these two training schemes for the top 4 best models in Table~\ref{tbl:ablation_analysis_architecture} (\ie, model 7, 8, 11, and 12). 
Results are shown in Figure~\ref{fig:ablation_analysis_loss}. 
Our proposed training scheme with InfoNCE loss is overall much better, it beats the BCE loss scheme on 14 out of total 16 cases. 
One reason is that BCE loss is very sensitive to the noises in the generated self-supervision data while InfoNCE loss is more robust to such label noise. 
Furthermore, we find that LBM matching can benefit more from our training scheme with InfoNCE loss -- with larger margin on all 8 cases, compared with the simple MLP matching.

\begin{figure}[!t]
  \centering
  \centerline{\includegraphics[width=0.48\textwidth]{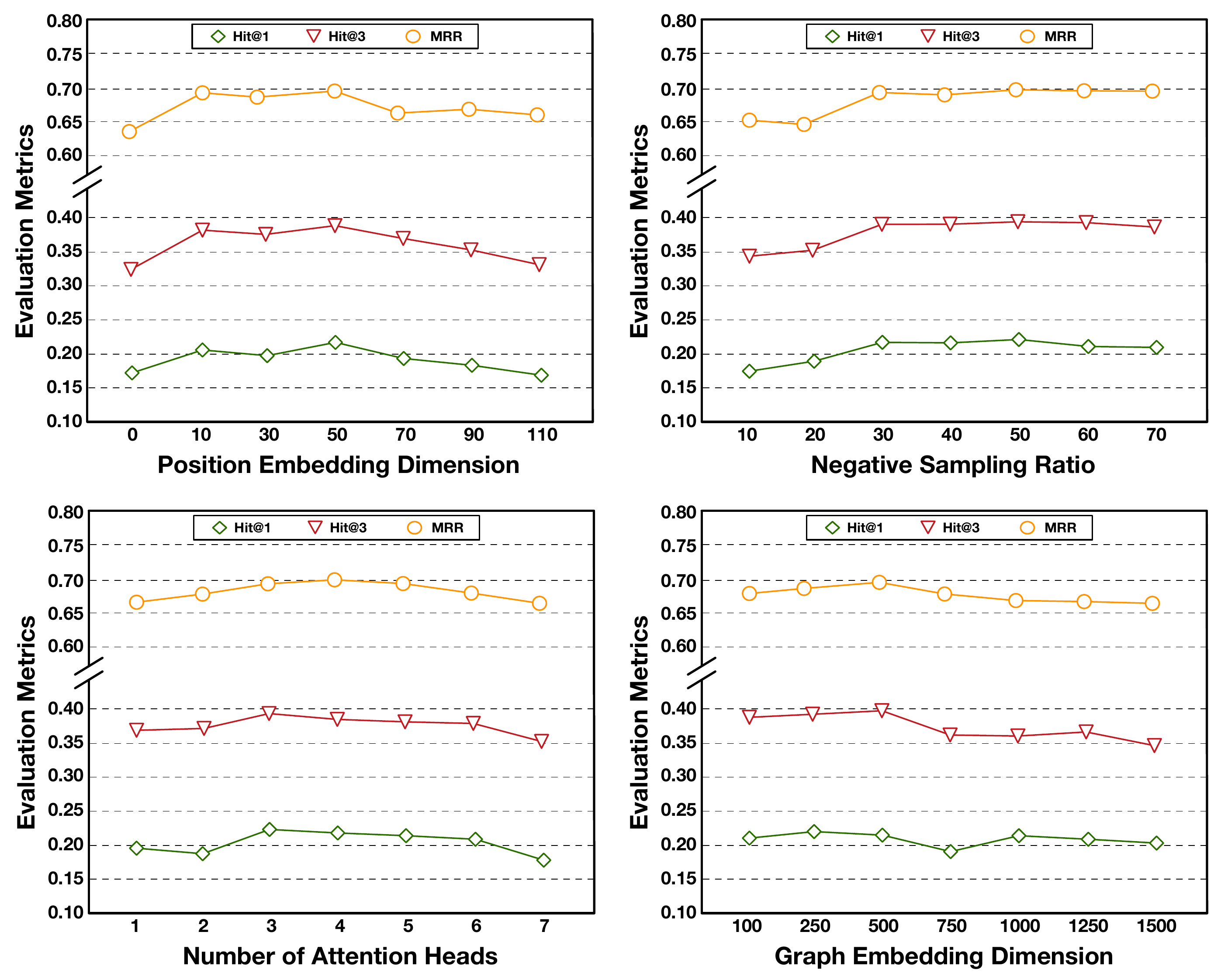}}
  \vspace{-0.2cm}
  \caption{Hyper-parameter sensitivity analysis on MAG-CS dataset. We use PGAT for graph propagation, WMR for graph readout, and LBM for query-graph matching. Model is trained using InfoNCE loss.}
  \label{fig:mag_parameter_sensitivity}
  \vspace{-0.3cm}
\end{figure}

\smallskip
\noindent \textbf{4. $~~$Hyper-parameter Sensitivity Analysis.}
We analyze how some hyper-parameters in \TaxoExpan affect the performance in Figure~\ref{fig:mag_parameter_sensitivity}.
First, we find that choosing an approximate position embedding dimension is important. 
The model performance increases as this dimensionality increases until it reaches about 50.
When we further increase position embedding dimension, the model will overfit and the performance decreases.
Second, we study the effect of negative sampling ratio $N$. 
As shown in Figure~\ref{fig:mag_parameter_sensitivity}, the model performance first increases as $N$ increases until it reaches about 30 and then becomes stable.
Finally, we examine two hyper-parameters controlling the model complexity: the number of heads in PGAT and the final graph embedding dimension.  
We observe that the best model performance is reached when the number of attention heads falls in range 3 to 5 and the graph embedding dimension is set to 500. 
Too many attention heads or too large graph embedding dimension will lead to overfit and performance degradation.

\begin{figure}[!t]
  \centering
  \centerline{\includegraphics[width=0.45\textwidth]{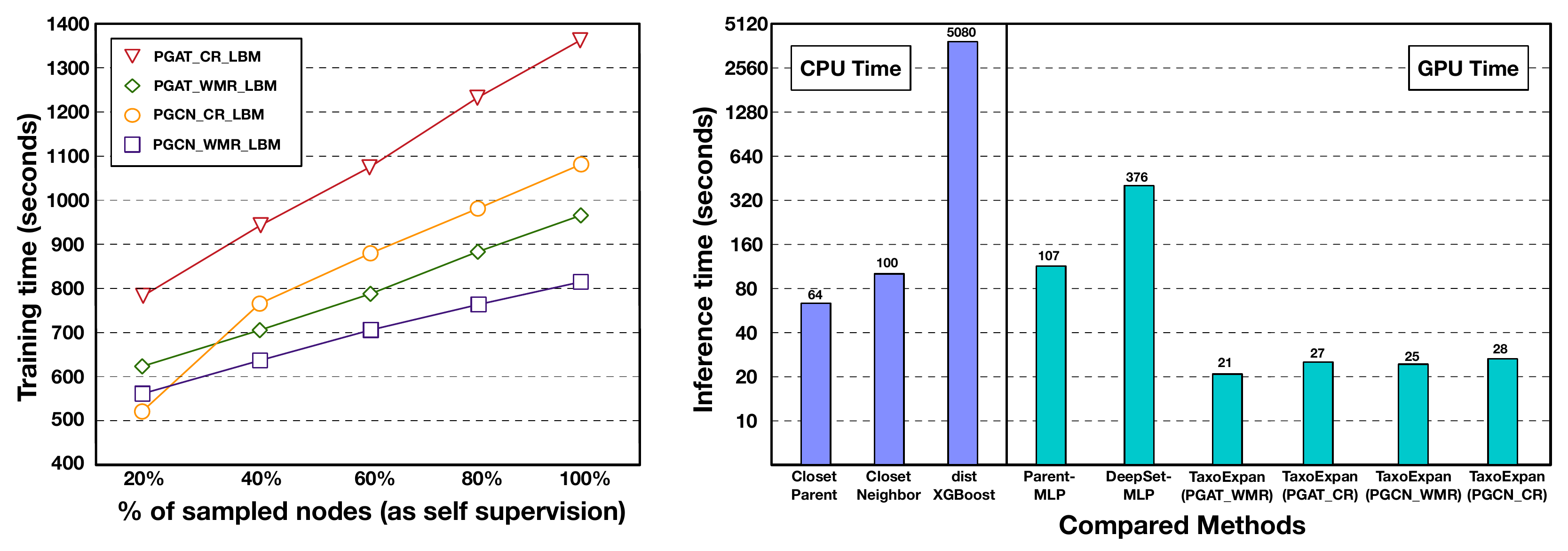}}
  \vspace{-0.2cm}
  \caption{(Left) Training time of 20 epochs on GPU with respect to \% of sampled nodes in the existing taxonomy. (Right) Inference time of all 2450 queries in MAG-CS dataset. Note here y-axis is in logarithm scale.}
  \label{fig:mag_scalability}
  \vspace{-0.2cm}
\end{figure}

\smallskip
\noindent \textbf{5. $~~$Efficiency and Scalability.}
We further analyze the scalability of \TaxoExpan and its efficiency during model inference stage. 
Figure~\ref{fig:mag_scalability} (left) tests the model scalability by running on MAG-CS dataset sampled using different ratios. 
The training time (of 20 epochs) are measured on one single K80 GPU. 
\TaxoExpan demonstrates a linear runtime trend, which validates our complexity analysis in Sect.~\ref{subsec:learning}. 
Second, Figure~\ref{fig:mag_scalability} (right) shows that \TaxoExpan is very efficient during model inference stage. 
Using GPU, \TaxoExpan takes less than 30 seconds to predict the anchor positions for all 2450 new query concepts. 

\begin{figure*}[!t]
  \centering
  \centerline{\includegraphics[width=1.0\textwidth]{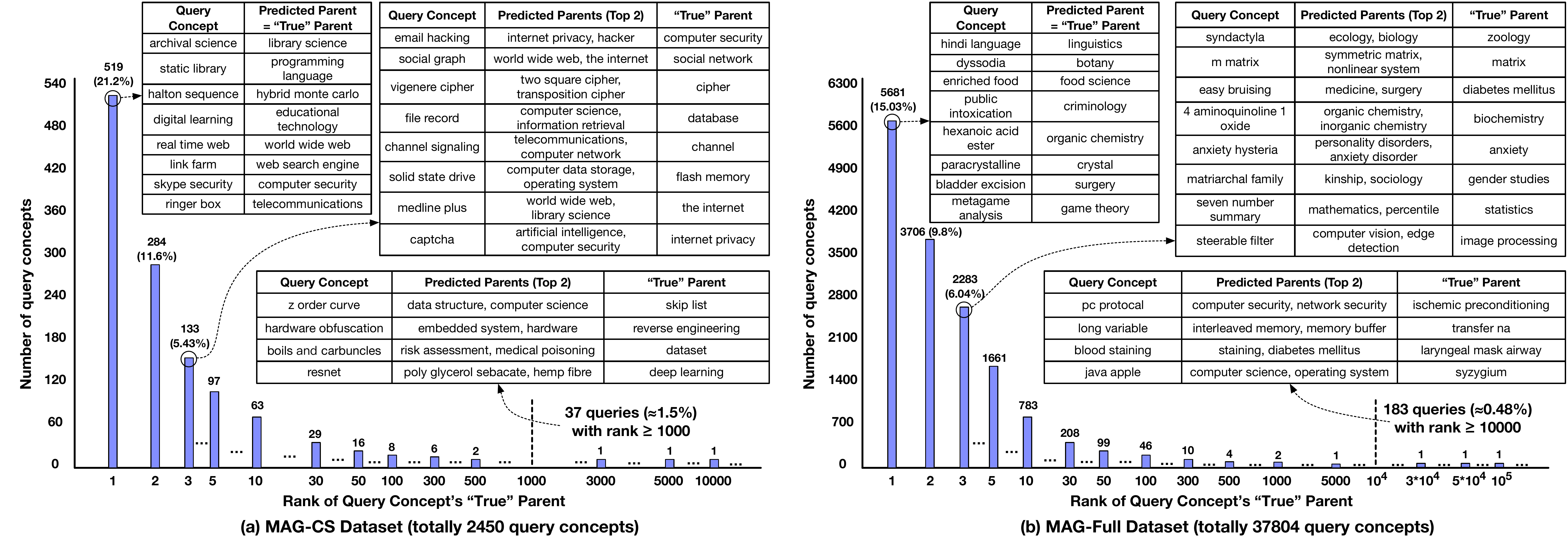}}
  \vspace{-0.01cm}
  \caption{Example output of \TaxoExpanBf on MAG-CS and MAG-Full datasets. We draw a histogram of the ranks of query concepts' true parents within the rank list returned by \TaxoExpanBf. In subfigure (a), for example, we have 519 (out of 2450) queries that their parents are exactly ranked in the first position.}
  \label{fig:case_study}
  \vspace{-0.01cm}
\end{figure*}

\smallskip
\noindent \textbf{6. $~~$Case Study.}
Figure~\ref{fig:case_study} shows some outputs of \TaxoExpan on both MAG-CS and MAG-Full datasets. 
On MAG-CS dataset, we can see that over 20\% of queries have their true parents correctly ranked at the first position and less than 1.5\% queries have their ``true'' parents ranked outside of top 1000 positions. 
Among these 1.5\% significantly wrong queries, we find some of them actually have incorrect existing parents.
For example, the concept \mquote{boils and carbuncles}, which is a disease entity, is mistakenly put under parent node \mquote{dataset}. 
Similar cases also happen on MAG-Full dataset where we find the concept \mquote{blood staining} is currently under \mquote{laryngeal mask airway}. 

Besides the above label errors, we also observe two common mistake patterns.
The first type of mistakes is caused by term ambiguity.
For instance, the term \mquote{java} in concept \mquote{java apple} refers to an island in Indonesia where fruit apple is produced, rather than a programming language used in Apple company. 
The second type of mistakes results from term granularity.
For example, \TaxoExpan outputs the two most likely parent nodes of concept \mquote{captcha} are \mquote{artificial intelligence} and \mquote{computer security}.
Although these two concepts are certainly relevant to \mquote{captcha}, they are too general compared to its true parent node \mquote{internet privacy}.

Finally, we observe that \TaxoExpan can return very sensible anchor positions of query concepts, even though they are not exactly the current ``true'' parents. 
For example, the concept \mquote{medline plus} refers to a large online medical library and thus is related to both \mquote{world wide web} and \mquote{library science}. 
Also, the concept \mquote{email hacking} is clearly relevant to both \mquote{internet privacy} and \mquote{hacker}. 

\smallskip
\noindent \textbf{7. \TaxoExpanBf for Taxonomy Self-Cleaning.}
From the above case studies, we find another interesting application of \TaxoExpan is to use it for cleaning the existing taxonomy. 
Specifically, we partition all leaf nodes of the existing taxonomy into 5 groups and randomly mask one group of nodes. 
Then, we train a \TaxoExpan model on the remaining nodes and predict on the masked leaf nodes. 
Next, we select those entities whose true parents appear at the bottom of the rank lists returned by \TaxoExpan (\ie, the long-tail part of two histograms in Figure~\ref{fig:case_study}). 
The parents of those selected entities are highly questionable and calls for further manual inspections. 
Our preliminary experiments on the MAG-CS taxonomy shows that about 30\% of these entities have existing parent nodes which are less appropriate than the parents inferred by \TaxoExpan. 

\subsection{Evaluation on SemEval Task Benchmark}

\subsubsection{Datasets}

We further evaluate \TaxoExpan using SemEval Task 14 Benchmark dataset\footnote{\scriptsize \url{http://alt.qcri.org/semeval2016/task14/}.} \cite{Jurgens2016SemEval2016T1} which includes WordNet 3.0 as the existing taxonomy and additional 1,000 domain-specific concepts with manual labels, split into 400 training concepts and 600 testing concepts. 
Each concept is either a verb or a noun and has a textual definition of a few sentences. 
The original task goal is to enrich the taxonomy by performing two actions for each new concept: (1) \emph{attach}, where a new concept is treated as a new synset and is attached as a hyponym of one existing synset in WordNet, and (2) \emph{merge}, where a new concept is merged into an existing synset. 
However, previous state-of-the-art methods \cite{Jurgens2016SemEval2016T1,Schlichtkrull2016MSejrKuAS,Vedula2018EnrichingTW}, including the winning solution, are only performing the \emph{attach} operation. 
In this work, we also follow this convention and attach each new concept to the top-ranked synset in the WordNet. 
Finally, we obtain the initial feature vectors (for both new concepts and existing words in the WordNet) using pre-trained subword-aware fasttext embeddings\footnote{\scriptsize We use the wiki-news-300d-1M-subword.vec.zip version on fastText official website.}. 
For each concept, we generate its definition embedding and name embedding by averaging the embedding of each token in its textual definition and name string, correspondingly. 
Then, we sum the definition and name embeddings of a concept and use them as the initial embeddings for the \TaxoExpan model.  

\subsubsection{Evaluation Metrics}
We use the three official metrics defined in original SemEval Task 14 for evaluation:
\begin{enumerate}[leftmargin=*]
\item \textbf{Accuracy (Wu\&P)} is the semantic similarity between a predicted parent node $x_{p}$ and the true parent $x_{t}$, calculated as $\text{Wu\&P}(x_{p}, x_{t}) = \frac{2\cdot depth_{LCA(x_p, x_t)}}{depth_{x_p} + depth_{x_t}}$, where $depth_{x}$ is the depth of node $x$ is the WordNet taxonomy and $LCA(x_p, x_t)$ represents the Least Common Ancestor of $x_p$ and $x_t$. 
\item \textbf{Recall} is the percentage of concepts for which an attached parent is predicted\footnote{\scriptsize This metric is used because the original task allows a model to decline to place new concepts in order to avoid making placements with low confidence.}. 
\item \textbf{F1} is the harmonic mean of Wu\&P accuracy and recall. 
\end{enumerate}
%

\subsubsection{Baseline Methods}
We compare the following methods:
\begin{enumerate}[leftmargin=*]
\item \textbf{FWFS} \cite{Jurgens2016SemEval2016T1}: The original baseline in Task 14. Given a concept $c$ with its definition $d_{c}$, this method picks the first word $w$ in $d_{c}$ that has the same part of speech as $c$ and treats this word as the parent node of $c$. 
\item \textbf{MSejrKU} \cite{Schlichtkrull2016MSejrKuAS}: The winning solution of Task 14. This method leverages distributional and syntactic features to train a SVM classifier which is then used to predict the goodness of fit for a new concept with an existing synset in WordNet. 
\item \textbf{ETF} \cite{Vedula2018EnrichingTW}: The current state-of-the-art method that learns a LambdaMART model with 15 manually designed features, including topological features from the taxonomy's graph structure and semantic features from corpus and Bing search results. 
\item \textbf{ETF-FWFS} \cite{Vedula2018EnrichingTW}: The ensemble model of FWFS and ETF, which adds the FWFS property as a binary feature into the LambdaMART model in ETF. 
\item \textbf{dist-XGBoost}: The same tree boosting model described in the previous subsection~\ref{subsec:compare_methods}.
\item \TaxoExpanBf: Our proposed taxonomy expansion framework.
\item \TaxoExpanFWFSBf: Similar to ETF-FWFS, this is the ensemble model of FWFS and \TaxoExpan. We treat the FWFS heuristic as a binary feature and add it into the final matching module.
\end{enumerate}
For all previous methods, we directly report their best performance in the literature.
For the remaining methods, we tune them following the same procedure described in the Section~\ref{subsubsec:implementation_details}.  

\begin{table}[!h]
	\centering
	\caption{Model performance on SemEval dataset. \TaxoExpanBf versus all previous state-of-the-art methods. We report the best performance of all existing methods in the literature.}
	 \label{tbl:overall_semeval_results}
	 \vspace{-0.05cm} 
	 \scalebox{1.05}{
	 	\begin{tabular}{c|ccc}
		\toprule
		\textbf{Method} & \textbf{Wu\&P} & \textbf{Recall} & \textbf{F1}  \\
		\midrule
		MSejrKU \cite{Schlichtkrull2016MSejrKuAS} & 0.523  & 0.973  & 0.680   \\
		FWFS \cite{Jurgens2016SemEval2016T1} & 0.514  & 1.000  & 0.679   \\
		ETF \cite{Vedula2018EnrichingTW} & 0.473  & 1.000  & 0.642   \\
		ETF-FWFS \cite{Vedula2018EnrichingTW} & 0.562  & 1.000  & 0.720   \\
		\midrule
		dist-XGBoost & 0.528  & 1.000  & 0.691   \\  
		\TaxoExpan & 0.543  & 1.000  & 0.704   \\  
		\TaxoExpanFWFS & 0.566  & 1.000  & 0.723   \\  
		\bottomrule
	\end{tabular}
 }
 \vspace{-0.05cm}
\end{table}

\subsubsection{Experimental Results}
Table~\ref{tbl:overall_semeval_results} shows the experimental results on SemEval dataset.
We can see that both dist-XGBoost and \TaxoExpan methods can outperform the previous winning system of this task (\ie, MSejrKU) and the baseline ETF. 
In addition, we can see the FWFS heuristic is indeed very powerful for this dataset and incorporating it as a strong feature can significantly boost the performance.
However, this feature requires human-labeled definition sentences and thus can not be easily generalized to taxonomies other than WordNet.
Finally, we show that \TaxoExpanFWFS can achieve the new state-of-the-art performance on this dataset.

\section{Conclusion}\label{sec:conclusion}
This paper studies taxonomy expansion when no human labeled supervision data are given. 
We propose a novel \TaxoExpan framework which generates self-supervision data from the existing taxonomy and learns a position-enhanced GNN model for expansion. 
To make the best use of self-supervision data, we design a noise-robust objective for effective model training.  
Extensive experiments demonstrate the effectiveness and efficiency of \TaxoExpan on three taxonomies from different domains. 
Interesting future work includes modeling inter-dependency among new concepts, leveraging current method to cleaning the input existing taxonomy, and incorporating feedbacks from downstream applications (\eg, search \& recommendation) to generate more diverse supervision signals for expanding the taxonomy.

\section{Acknowledgement}\label{sec:ack}
Research was sponsored in part by DARPA under Agreements No. W911NF-17-C-0099 and FA8750-19-2-1004, National Science Foundation IIS 16-18481, IIS 17-04532, and IIS-17-41317, and DTRA HDTRA11810026. Any opinions, findings, and conclusions or recommendations expressed in this document are those of the author(s) and should not be interpreted as the views of any U.S. Government. 
We thank Yuxiao Dong, Ziniu Hu, Li Ma for insightful discussions on this project and anonymous reviewers for valuable feedbacks.

\section*{Appendix}\label{appendix}

\subsection*{Proof of Loss Function}
Here we prove that optimizing the loss function in Eq. (\ref{eq:loss}) will result in $f(\cdot)$ estimating the probability density in Eq. (\ref{eq:density}).
By construction, $\mathbf{X}$ contains query $n_c$'s one positive anchor (\ie, its true parent $n_p$) sampled from the true distribution $\P(a_i|n_c)$ and $N$ negative anchors $\{ n_{r}^{l} |_{l=1}^{N}\}$ sampled from a uniform distribution $\P(a_i)$. 
If we merge these $N+1$ anchors into a small set and consider the task of selecting true anchor $n_p$'s position $j^{*}$ in $[1, 2, \dots, N+1]$, we can view Eq. (\ref{eq:loss}) as the cross entropy of position distribution $\hat{\P}$ from model prediction relative to the true distribution $\P^{*}$.
Specifically, the model predicted position distribution $\hat{\P}_{j} = \frac{f(a_j, n_c)}{\sum_{k=1}^{N+1} f(a_k, n_c)}$ where one of $\{a_{k} | _{k=1}^{N+1} \}$ is the true anchor and all the others are negative anchors. 
Meanwhile, in the true position distribution: 
\begin{displaymath}
\small
\P^{*}_j = \frac{\P(a_j | n_c)\prod_{l \neq j}\P(a_l)}{\sum_{k =1}^{N+1} \left(\P(a_{k}|n_c) \prod_{l \neq k}\P(a_l)\right) }  = \frac{ \frac{\P(a_j | n_c)}{\P(a_j)}}{\sum_{k =1}^{N+1}  \frac{\P(a_{k} | n_c)}{\P(a_{k})} }.
\end{displaymath}
From above, we can see that the optimal value for $f(a_j, n_c)$ is proportional to $\frac{\P(a_j | n_c)}{\P(a_j)}$. 


\bibliographystyle{ACM-Reference-Format}
\bibliography{cited}

\end{document}